\documentclass[letterpaper, 10 pt, conference]{ieeeconf}  

\IEEEoverridecommandlockouts                              

\usepackage[absolute,overlay]{textpos}
\usepackage{xcolor}

\newcommand{\norm}[1]{\left\lVert#1\right\rVert}

\usepackage{amsmath} 
\usepackage{amssymb}  
\usepackage{amsmath}
\usepackage{booktabs}

\title{\LARGE \bf
Time-to-Label: Temporal Consistency for Self-Supervised Monocular 3D Object Detection
}
\author{Issa Mouawad$^{1}$, Nikolas Brasch$^{2}$, Fabian Manhardt$^{3}$,  Federico Tombari$^{2,3}$, Francesca Odone$^{1}$
\thanks{$^1$:MaLGa-DIBRIS, University of Genoa, Italy,
{\tt\small issa.mouawad@edu.unige.it, francesca.odone@unige.it}}
\thanks{$^2$:Technical University of Munich, Germany; {\tt\small \{nikolas.brasch, federico.tombari\}@tum.de}}
\thanks{$^3$:Google Inc. {\tt\small fabianmanhardt@google.com}}}

\begin{document}
\begin{textblock*}{20cm}(0.5cm,0.5cm) 
   This work has been submitted to the IEEE for possible publication. Copyright may be transferred without notice, after which this version may no longer be accessible.
\end{textblock*}
\maketitle
\thispagestyle{empty}
\pagestyle{empty}

\begin{abstract}
Monocular 3D object detection continues to attract attention due to the cost benefits and wider availability of RGB cameras. Despite the recent advances and the ability to acquire data at scale, annotation cost and complexity still limit the size of 3D object detection datasets in the supervised settings. Self-supervised methods, on the other hand, aim at training deep networks relying on pretext tasks or various consistency constraints. Moreover, other 3D perception tasks (such as depth estimation) have shown the benefits of temporal priors as a self-supervision signal. In this work, we argue that the temporal consistency on the level of object poses, provides an important supervision signal given the strong prior on physical motion. Specifically, we propose a self-supervised loss which uses this consistency, in addition to render-and-compare losses, to refine noisy pose predictions and derive high-quality pseudo labels. To assess the effectiveness of the proposed method, we finetune a synthetically trained monocular 3D object detection model using the pseudo-labels that we generated on real data. Evaluation on the standard KITTI3D benchmark demonstrates that our method reaches competitive performance compared to other monocular self-supervised and supervised methods.
\end{abstract}

\section{Introduction}
\label{sec:intro}

Enabling computer systems to perceive the 3D world has received a lot of attention in the recent years, as it serves numerous applications in autonomous driving, robotics and many more~\cite{roi10d, 3drcnn, wang2021demograsp, tateno2017cnn}. In particular, monocular approaches are growing in interest due to the advantages in availability and cost of RGB cameras, in comparison with actual 3D sensors. As for autonomous driving, lidar sensors are currently part of the standard equipment of every car, nevertheless, due to cost and availability advantages, in addition to the higher spatial resolution, cameras will most likely be the main sensor in future large-scale deployment. Hence, being able to perceive the 3D scene and detect all objects within it from RGB data alone is of high importance.

Unfortunately, the training of such models requires access to large-scale labeled datasets in order to meet generalization and accuracy standards. This is particularly challenging for 3D perception tasks central to  autonomous driving and robotics applications, as it requires to tightly hand-label millions of 3D bounding boxes. Nonetheless, it is worth mentioning that, although labeling such samples is very expensive, recording of appropriate data is fairly cheap and can be achieved by simply driving around. Therefore, being able to train on this data in a self-supervised fashion without the utilization of any labels would be highly beneficial, enabling to easily scale to larger volumes of training data and effectively pushing forward both accuracy and generalizabilty. While self-supervision has already been well explored in several disciplines~\cite{ssl1,ssl2}, only a handful of methods started very recently to apply self-supervision to the domain of 3D object detection~\cite{self6d,wang2021occlusion,autolabeling}. 
These very recent works commonly leverage 3D shape priors and differentiable rendering pipelines to derive consistency losses as means of supervision, in the absence of annotations~\cite{autolabeling,self6d,koestler}. While these methods generally perform well, they completely neglect any temporal information, which can however serve as a very strong source of 3D supervision as demonstrated in many works from other domains~\cite{godard,pillai2019superdepth}.

In this work we propose a novel pipeline that leverages raw lidar information together with temporal information to establish supervision for monocular 3D object detection without the need for any real pose labels. In particular, we train a state-of-the-art monocular 3D object detector completely in simulation. In the following step, we utilize the obtained model to label real data with pseudo labels. These pseudo labels are then refined based on establishing coherence with the 3D scene, in the form of a 3D lidar point cloud, and temporal information from neighboring frames.

\noindent Our contributions can be summarized as follows. We propose

\begin{itemize}
    \item a simple, yet effective, \emph{motion classification module} which can estimate the motion state despite noise and outliers in the pose estimates,
    \item a novel \emph{temporal consistency loss}, which builds on the temporal prior of objects trajectories and helps accurately refining noisy pose estimates,
    \item a self-supervised temporally-aware framework for generating 3D labels from image sequences and un-annotated lidar pointclouds.
   
\end{itemize}
  \begin{figure*}
\def\svgwidth{\textwidth}
\centering
      \includegraphics[height=7cm]{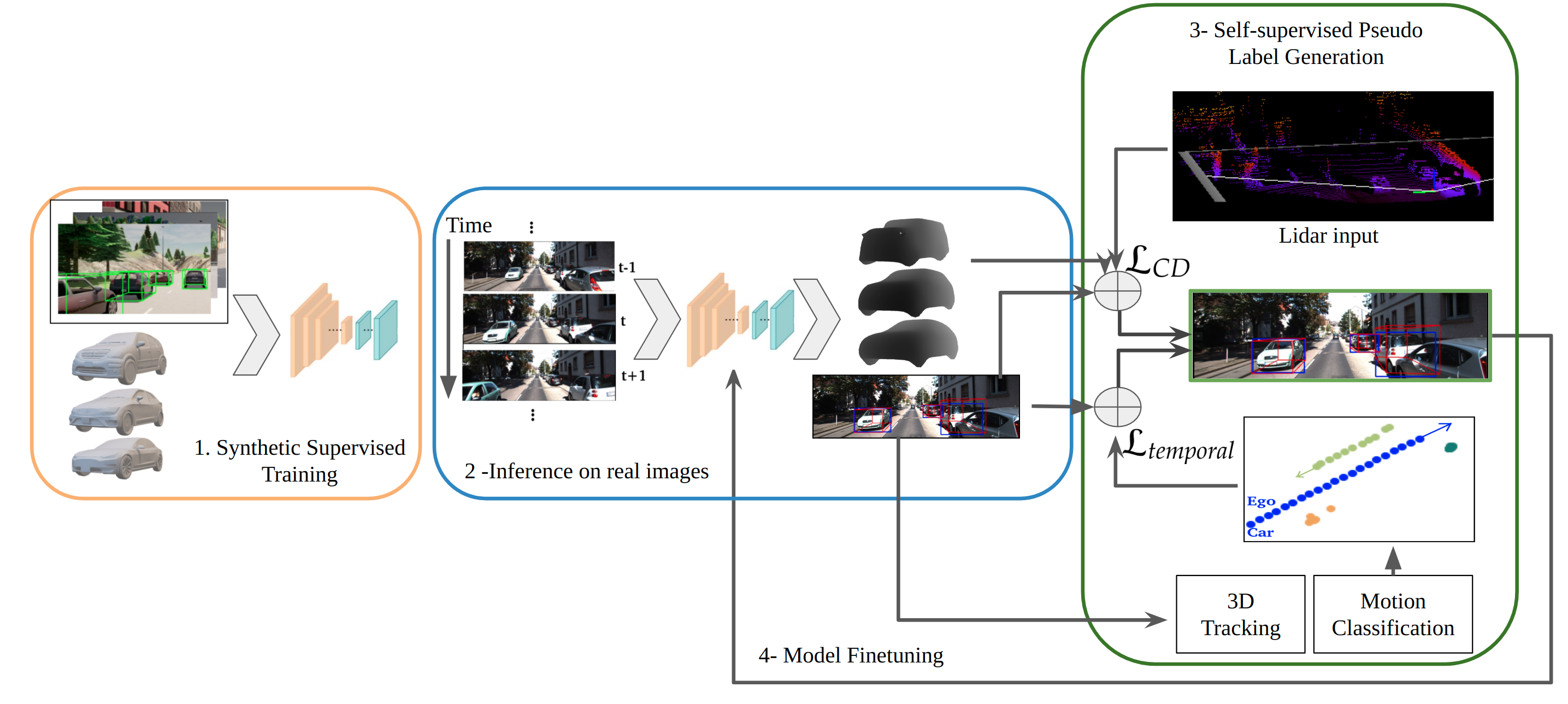}
      
      \caption{Schematic overview of our proposed method. (1) We first train a monocular 3D object detector completely in simulation in a supervised fashion. (2-4) Using the 3D scene geometry together with temporal data, we create high-quality pseudo labels that we use to finetune the model.}
      \label{fig:method}
  \end{figure*}

\section{Related Work}
\subsection{Monocular 3D Object Detection}
Research in monocular 3D object detection has seen a significant surge recently. Recovering 6D pose of known objects, or, more recently, predicting rough 3D shapes (cuboids) in a category-based detection, represents an essential building block in 3D perception systems, for applications in Robotics, and Autonomous Driving, among others.

According to a recent taxonomy \cite{3dsurvey}, monocular 3D object detection methods can be grouped into methods which lift 2D detections to 3D (referred to as result lifting-based methods) and methods which process 3D features derived from the 2D image plane.

Early methods are dominantly from the first category.
Such works first estimate 2D properties such as the 2D location, dimensions and orientation. Then, the 3D center is retrieved by estimating the center's depth z and then back-projecting the 2D center according to the regressed depth to 3D. 
Several of those detectors are built on top of region-proposal 2D detectors \cite{faster} and use the RPN features to estimate 3D properties of objects \cite{mono3d,3dop,roi10d,mousavian20173d}.
While others exploit the low computational cost of single-stage 2D detectors to deliver an efficient 3D inference \cite{monoflex,monodis,m3drpn,movi3d,rtm3d}.

Other recent methods, falling in the second category, propose to handle 3D aspects early in the pipeline such as pseudo-lidar methods \cite{pseudolidar} which transform depth to a lidar point cloud, and methods which directly transform the CNN features from the 2D image plane to the world 3D reference system (or directly to the bird's eye view) \cite{philion2020lift}.

Despite the recent progress of monocular methods, learning 3D object detection from images is an ill-posed problem and requires access to a large number of examples. In a supervised setting, manually produced labels require an expensive and tedious procedure to annotate. On the other hand, using a limited amount of labeled data carries the risks of over-fitting and poor performance. 
\subsection{Self-Supervised Learning}
In order to exploit the potential of large-scale data,  learning without the need for annotations has gained a lot of attention recently.
Self-supervised representation learning (SSRL) \cite{ssrl} is used to extract meaningful representations for solving various visual tasks, such as image classification \cite{rotation_ssrl}, object detection \cite{jigsaw}, and visual tracking \cite{tracking_temporal}, to name a few. 
Furthermore, self-supervised learning has been used to directly train models on the downstream task without supervision, harnessing additional geometric and consistency priors. This direction has witnessed a growing interest lately with methods addressing monocular depth estimation \cite{depth1,depth2}, optical flow \cite{ssl_opflow}, and scene flow prediction \cite{ssl_sceneflow}.

In 3D object detection, the self-supervision typically requires a strong prior on the scene, which is commonly encoded in the form of initial predictions generated by a pre-trained  model \cite{self6d}, or a latent variable decoders \cite{monodr} in addition to 3D shape object priors \cite{deepsdf}. Exploiting Differentiable Learning pipelines \cite{drsurvey}, and render-and-compare losses \cite{3drcnn}, different consistency losses are proposed both in 3D  \cite{monodr,autolabeling,koestler} and in 2D using mask, appearance and other photometric similarities \cite{koestler,monodr,self6d}. To improve learning stability and provide fairer comparisons, a recent direction frames the self-supervision as a pseudo-label generation. In this framework, pseudo-labels are generated using initial hypotheses of 6D poses, these labels are then refined and subsequently utilized to train a 3D object detection network in a supervised manner. Recent works either use a combination of RGB images and lidar point cloud during training \cite{autolabeling}, or rely solely on lidar point clouds to generate 3D pseudo-labels \cite{lpcg,st3d,caine2021pseudo}.

Nonetheless, although achieving great results despite the lack of labels, none of these approaches make use of temporal consistency across frames as a prior on
object pose estimation to further strengthen the pesudo label generation.

\subsection{Temporal Consistency for Self-supervision}
Temporal consistency across time in videos represents a rich source of prior information. Recently, several visual tasks try to explicitly model and exploit this consistency. In visual tracking \cite{tracking_temporal} temporal consistency is used to self-supervise visual correspondence representation learning.
In tasks where time is a key component, such as scene flow, a good temporal prior is essential to capture meaningful representations \cite{sceneflow}. Notice that also pixel-wise tasks, not directly related to time evolution, can benefit from temporal consistency both during training and inference, as for example depth estimation \cite{depth1,depth2} and semantic segmentation \cite{semseg}.
 
Temporal Consistency remains a largely unexplored direction for 3D object detection supervision, despite recent attempts to use it in a supervised framework \cite{4d} or in domain adaptation for lidar object detection \cite{sfuda}.
  
 \section{Methodology}

We aim to train a monocular 3D object detector from unlabelled data sequences. Unfortunately, this is a very challenging and highly ill-posed problem due to the nature of the perspective projection onto the image plane. Therefore, we slightly relax the problem, assuming to have access to a lidar and IMU/GPS sensor at training time. We believe that this is a fair assumption as almost all common benchmark datasets or test vehicles provide the respective data. Notice that at inference time, our network is capable of estimating the 6D object pose from a single RGB image alone. 
 
 Inspired by~\cite{self6d, wang2021occlusion}, we propose to pre-train a 6D pose and 3D shape prediction network on synthetic images (step 1 in Figure \ref{fig:method}).
 Next, using the obtained trained base model, we can estimate initial hypotheses of the object pose on real data (step 2 in Figure \ref{fig:method}). Furthermore, leveraging scene geometry and \textit{temporal consistency}, we generate temporally-aware pseudo-labels (step 3), which can be harnessed to finetune the base model (step 4). Steps (2 to 4) are then iteratively repeated to further improve the pseudo-labels quality and thus the finetuned model, as discussed in Section \ref{pseudo_label_generation}. 

\subsection{Training in Simulation}\label{sec:base}

\begin{figure}

\centering
      \includegraphics[height=5cm]{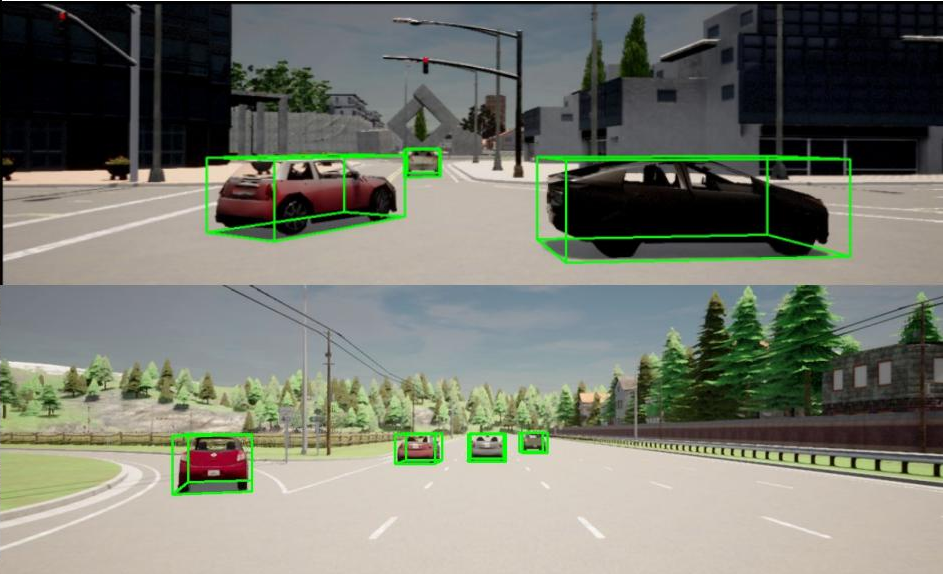}
      
      \caption{Training samples generated with the CARLA simulator~\cite{carla}}
      \label{fig:carla_samples}
  \end{figure}

As our self-supervision requires 3D shapes to enforce consistency with the 3D scene, we require the monocular 3D detector to predict additional object properties to encode the 3D shape.
To this end, we utilize CAD models of various cars brands to obtain a low dimensional shape space with PCA. We then fully supervise the whole architecture (3D detection and 3D shape prediction) on synthetic data from the open-source CARLA~\cite{carla} simulator adding an L2 loss term for shape supervision in the latent embedding of the PCA space. At inference, we can use the estimated latent encoding  to reconstruct the underlying 3D shape \textit{w.r.t.} the learned principal components.  Exemplary training samples can be found in Figure \ref{fig:carla_samples}.

\subsection{Generation of Initial Pose Labels}\label{sec:pseudo_labels}
After having trained our base model in simulation, we feed it with real RGB images to predict the 3D translation $t_i$, the rotation around the vertical axis $yaw_i$, and the shape embedding $s_i$ for each object $i$. We then reconstruct the corresponding 3D point cloud $\hat{P_i} = D_{PCA}(s_i)$ using the learned principal components. To further improve the obtained pose labels, we apply an additional refinement step using lidar information. Therefore, we transform the estimated pointcloud to the lidar reference frame following
\begin{equation}
     \hat{P}_{lidar} := (R_{yaw_i} \hat{P_i} + t_i),
     \label{eqn:trans}
\end{equation}
with $R_{yaw_i}$ referring to the 3D rotation matrix w.r.t the angle $yaw_i$.
Thereafter, we employ Chamfer distance \cite{chamfer} to measure the misalignment between the observed lidar scan  $\widetilde{P}$ and the predicted point cloud $\hat{P}_{lidar}$ as follows:
\begin{equation} \label{eq:cd}
    \mathcal{L}_{CD} = \sum_{x\in \widetilde{P}} \min_{y \in \hat{P}_{lidar}} ||x-y||^2_2 + \sum_{y\in \hat{P}_{lidar}} \min_{x \in \widetilde{P}} ||x-y||^2_2,
\end{equation}
which we optimize for our pose parameters $R_i$ and $t_i$ to improve their fit with the scene geometry.

As the observed lidar scan $\widetilde{P}$ is obtained for the whole scene, we need to filter-out points which do not belong to the object. This is accomplished by means of removing all points whose projection do not belong to the instance mask $M_i$ as predicted by Mask R-CNN~\cite{mask}. Note that our Mask R-CNN is pre-trained on a general purpose dataset (\textit{i.e.} COCO~\cite{coco}), thus, requiring no domain-specific supervision.

\subsection{Supervision by Means of Temporal Consistency}\label{sec:temporal_consistency}
Due to several sources of errors such as low lidar-coverage (due to occlusion for example) or weak initial pose estimates, the supervision from 3D points alone often generates a noisy signal. Therefore, we strengthen the refinement using temporal sequences. Thereby, we build a unified and temporally coherent world reference system across each video sequence to handle the presence of ego-motion, as well as the motion of all other cars in the scene.

Nevertheless, to build such world-reference system we first need to understand our own motion, which we obtain from IMU/GPS measurements. Secondly, we are also required to understand the motion of each other vehicle. To this end, we simply utilize the online 3D tracker \cite{baseline_tracker} to build 3D trajectories from individual 3D predictions for each image in the sequence. Finally, to bring all objects to our temporally coherent world reference system, we project all observations $t_i$ to the camera system of the initial frame. The respective 3D translation  $t_{global_i}$ is obtained by multiplying the observation $t_i$ with $T_{i,0}$, the transformation from the reference system at frame i to that of frame 0, obtained from the IMU/GPS measurements, according to:
\begin{equation}
    t_{global_i} = T_{i,0} \cdot t_i.
\end{equation}

To bring the rotation into our world reference system $yaw_{global_i}$, we  compute the object's heading direction as
\begin{equation}
    P_{heading} = Rot_{yaw_i}[1,0,0]^T
\end{equation}
We then project $P_{heading}$ to the global reference system and calculate the angle of the vector $\overrightarrow{P_{heading}}$ with the horizontal axis, and employ it as the global object orientation.

We observe that a motion prior can provide an additional source of supervision which, depending on whether a car is parked or moving, can help  recovering noisy initial estimates. 
 Thereby we distinguish between two cases: (i) objects that are \emph{static} relative to the scene, (ii) objects that are \emph{moving}, while for objects which have less observations than $min\_frames = 6$, we do not apply the temporal consistency. Unfortunately, due to noisy estimates from the trained base model, retrieving the motion state is not straightforward. Yet, we argue that 
temporal evolution of the object position, can provide insights about a plausible motion pattern or noise caused by pose estimation. Therefore, we employ velocity profiles to perform the motion state classification.  First, we separately sum the instantaneous signed velocities across the X and Y axis. Moving objects yield large values, while static objects exhibit velocities with different signs, which cancel out and yield a small traversed distance. We use an empirical threshold of $3$ meters (computed from the average vehicle dimension) to obtain the motion state. Thereby, whenever the traversed distance is less than the threshold we consider the object as static and moving otherwise. While we deem the previous condition sufficient to classify an object as static, we observe some situations where few outlier observations cause static objects to produce larger velocity values. Thus, for objects which fail the first condition, we further check zero-crossings of the velocity profile since the frequency of zero-crossings is not affected by the actual amplitude of the velocity values. We then classify as static, objects where zero-crossings events happen in at least 40\% of the observations.
\begin{figure}

\centering
      \includegraphics[height=6.2cm]{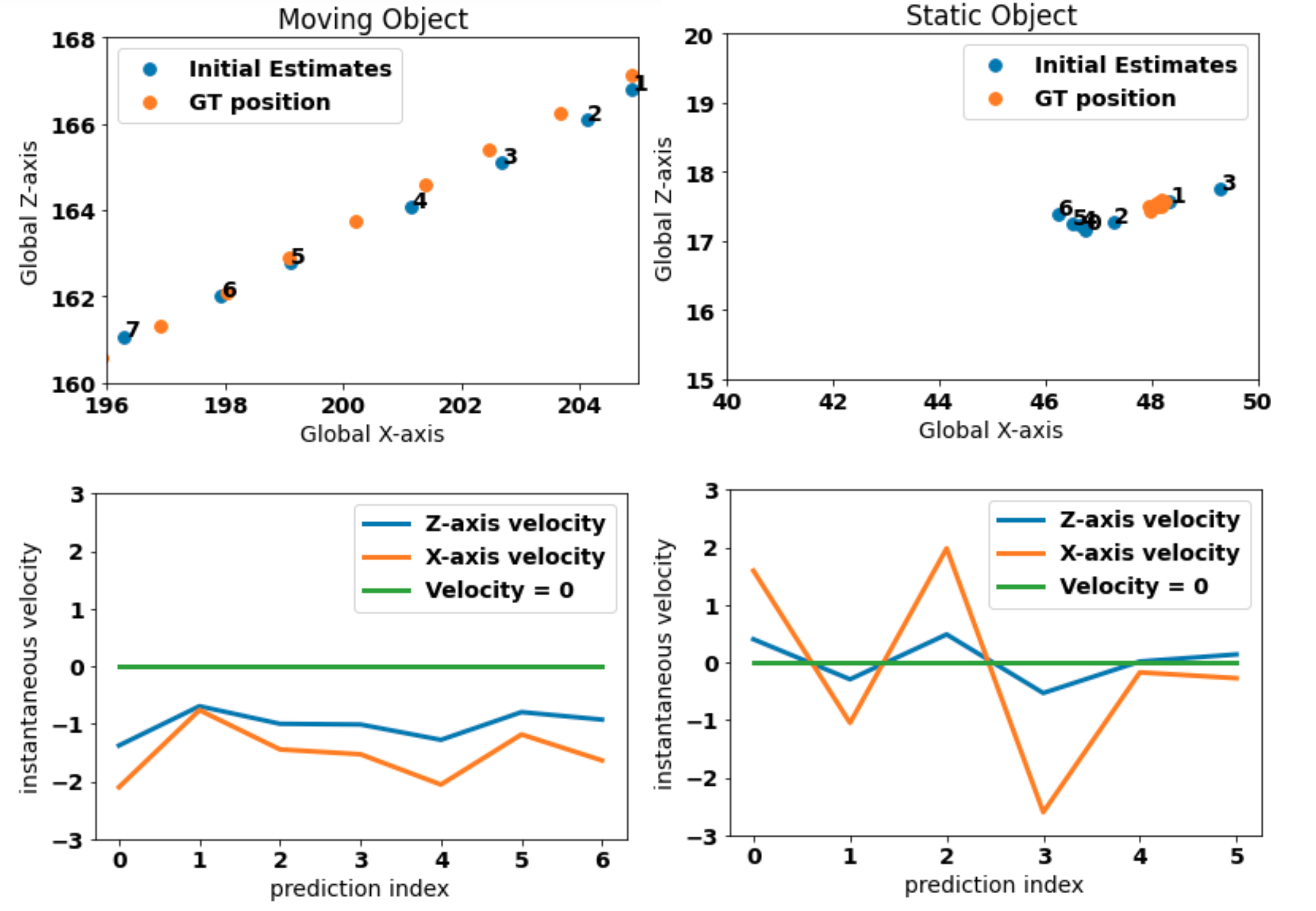}
      
      \caption{World reference system observations of a static object (right) and a moving one (left) with the corresponding velocity profiles (bottom)}
      \label{fig:motion_class}
  \end{figure}

Eventually, to enforce temporal consistency as supervision, we utilize two different formulations of our temporal consistency loss, depending on the motion state of the object. As for \textbf{static} objects, we first calculate the median scene position of all the observations $t_{glob}^{median}$, and consider it as an additional regularization for the refinement procedure (in addition to the Chamfer distance). Further, for the yaw angle of static cars, we express the observed yaw angles (of all the object instances) in our world reference system and then organize the angles in a 32-bin histogram. We pick the most frequent bin $yaw_{glob}^{mean}$ as the best estimate of the yaw angle to regularize the rotation, combining measurements from different view points.
Finally, to refine an observation $i$ of a static object, we optimize the following loss function
\begin{equation} \label{eq:loss_static}
\begin{split}
    \mathcal{L}_{static} &= \lambda_t \norm{t_i - t_i^{median}}_2^2\\
    &+ \lambda_r \norm{yaw_i - yaw_i^{mean}}_2^2,
\end{split}
\end{equation}
where the translation $t_i^{median}$ and the rotation $yaw_i^{mean}$ are both expressed in the local reference system by projecting the median pose back to the respective time frame. Notice that given the increased stability of the optimized pose, we further propagate it to adjacent frames (in which the object is not detected) to account for false negatives induced by large distance or truncation.
 
As for \textbf{moving} cars, we model the motion of a vehicle using a piece-wise linear trajectory (in the global reference system), which is a simple, yet expressive model for motion (assuming adequate video frame-rate of at least 10 fps).
We set the length of each segment to 10 frames, and fit a linear function using RANSAC. For each observation in the segment, we project the initial estimate on the RANSAC fitted motion model and obtain $t_{smoothed}$ as a regularization for the translation. We further assume the orientation to be roughly constant within the small time window and derive $yaw_{line}$ as the direction of the fitted line, which we then use as a regularization for the rotation.  This explicitly improves the orientation continuity and suppresses observation noise. Similar to Equation~\ref{eq:loss_static}, the loss function to optimize a moving vehicle observation is formulated as:
\begin{equation} \label{eq:loss_final}
\begin{split}
    \mathcal{L}_{moving} &=  \lambda_t \norm{t_i - t_{smoothed}}_2^2\\
    &+ \lambda_r \norm{yaw_i - yaw_{line}}_2^2.
\end{split}
\end{equation}

Summarizing, after training our model fully supervised in simulation using scenes generated from CARLA, we use it to compute noisy pseudo-labels for each frame. We then estimate the motion state of each vehicle and optimize the noisy observations accordingly to
\begin{multline*} \label{eq:loss_final}
    \mathcal{L} =  \mathcal{L}_{CD} + \mathcal{L}_{temporal}\\ \mathcal{L}_{temporal} =
    \begin{cases}
            \mathcal{L}_{static}, &         \text{if  static object}\\
            \mathcal{L}_{moving}, &         \text{if moving object}.
    \end{cases}
\end{multline*}
Afterwards, we finetune our initial model on the obtained pseudo-labels using the losses from \cite{monoflex}. This process is then repeated until convergence.
 
\section{Experiments}
In the following we will first present our evaluation protocol before we demonstrate several ablations, proving the usefulness of our contributions. We then conclude by comparing our method to several state-of-the-art works for training with and without real pose labels.

\subsection{Evaluation Protocol}
\paragraph{Datasets} We use the \textbf{KITTI} dataset, which is composed of a number of driving sequences with the corresponding lidar scans, both annotated with object bounding boxes. To learn 3D object detection in a self-supervised manner, we do not make use of any ground truth information during training, and we exclusively use them for evaluation purposes.
Specifically, we use the split proposed by \cite{split} to train our model without ground truth labels, while we use the validation split, to report results and compare with other methods both in terms of average precision in 3D (AP 3D) and for the boxes projected on the ground plane or AP bird's eye view (AP BEV).

\subsection{Synthetic Data Training}
We generate an urban dataset with Carla Simulator \cite{carla} having camera viewpoint and intrinsics similar to those in the KITTI dataset, comprising 50K images with 6D pose labels. We use 12 different car models in the training set, for which we generate PCA shape encodings.

As the choice of the backbone 3D object detector is not part of our main contribution, we simply adopt the state-of-the-art MonoFlex~\cite{monoflex} detector for all of our experiments. MonoFlex is built on top of CenterNet~\cite{centernet} and conducts single-stage anchor-free 2D object detection together with the estimation of object-wise properties, including 3D translation and horizontal plane rotation (yaw). Notice that MonoFlex only predicts one angle of the 3D rotation as it assumes that all objects stand on the ground plane, which is a common practice in the autonomous driving scenario. As mentioned in Section \ref{sec:intro}, we extend MonoFlex to predict 3D shape encodings supervised by the encodings of the ground truth CAD models. We further employ DLA-34 \cite{dla} as backbone initialized with Imagenet weights. After freezing the first three blocks of the backbone,  we train the rest of the network for 10 epochs on the tasks of interest, \textit{i.e.}  2D and 3D detection, and shape encoding regression. During training, we incorporate horizontal flip and color jitter augmentations, to minimize overfitting on the synthetic domain.

\subsection{Pseudo-label Generation}
\label{pseudo_label_generation}
We run pseudo-label generation routine for three iterations (using $\lambda_t = 0.25$ and $\lambda_r = 2$) and for 100 pose refinement steps, where we alternate between generating pseudo-labels and finetuning of the initial model on those pseudo-labels. 
When finetuning the model on the pseudo-labels we freeze an additional block in the backbone and we finetune all the heads as before, training for 4 epochs during each iteration.

\begin{figure*}
\def\svgwidth{\textwidth}
\centering
      \includegraphics[height=7cm]{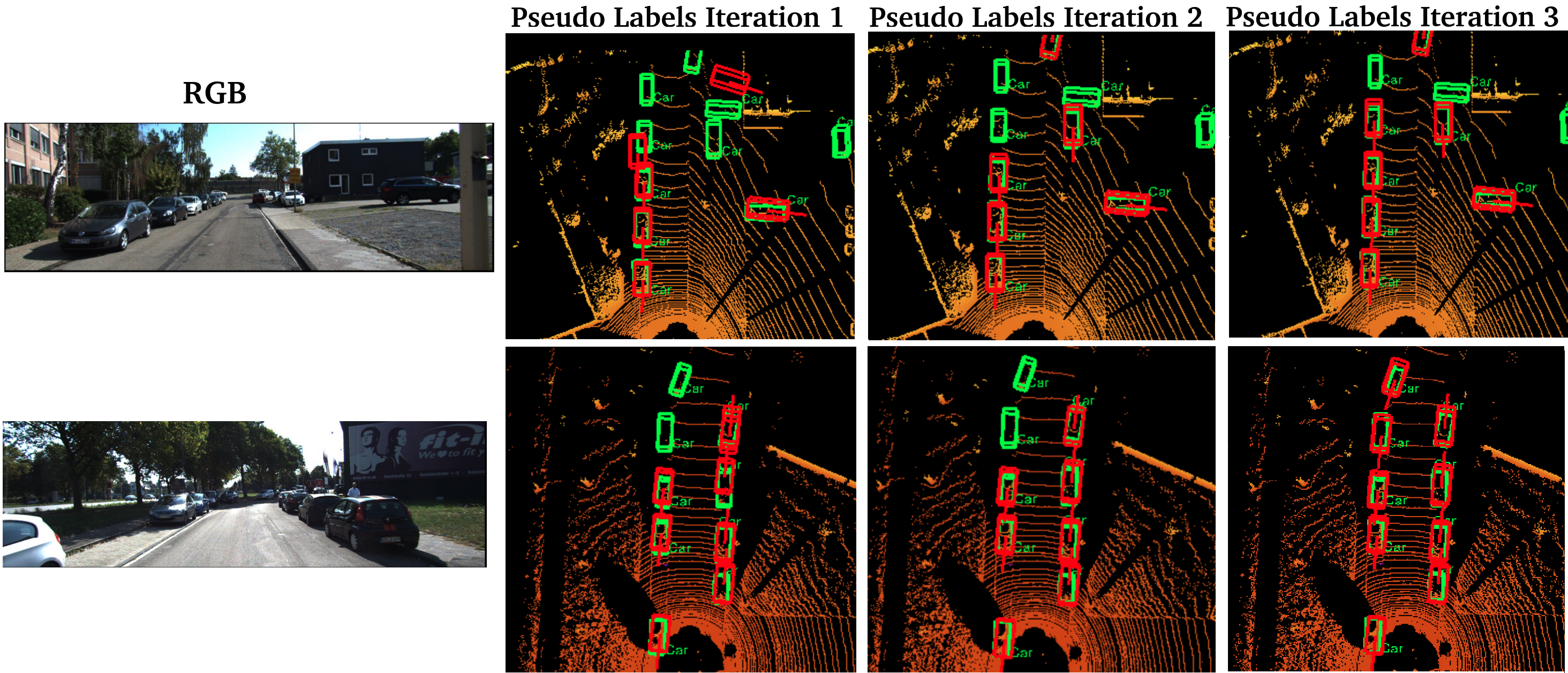}
      
      \caption{Pseudo-Labels (red boxes) qualitative assessment against the ground truth (green boxes) across different iterations (best viewed in color)}
      \label{fig:pseudo_qual}
  \end{figure*}

\begin{figure*}
\def\svgwidth{\textwidth}
\centering
      \includegraphics[height=8cm]{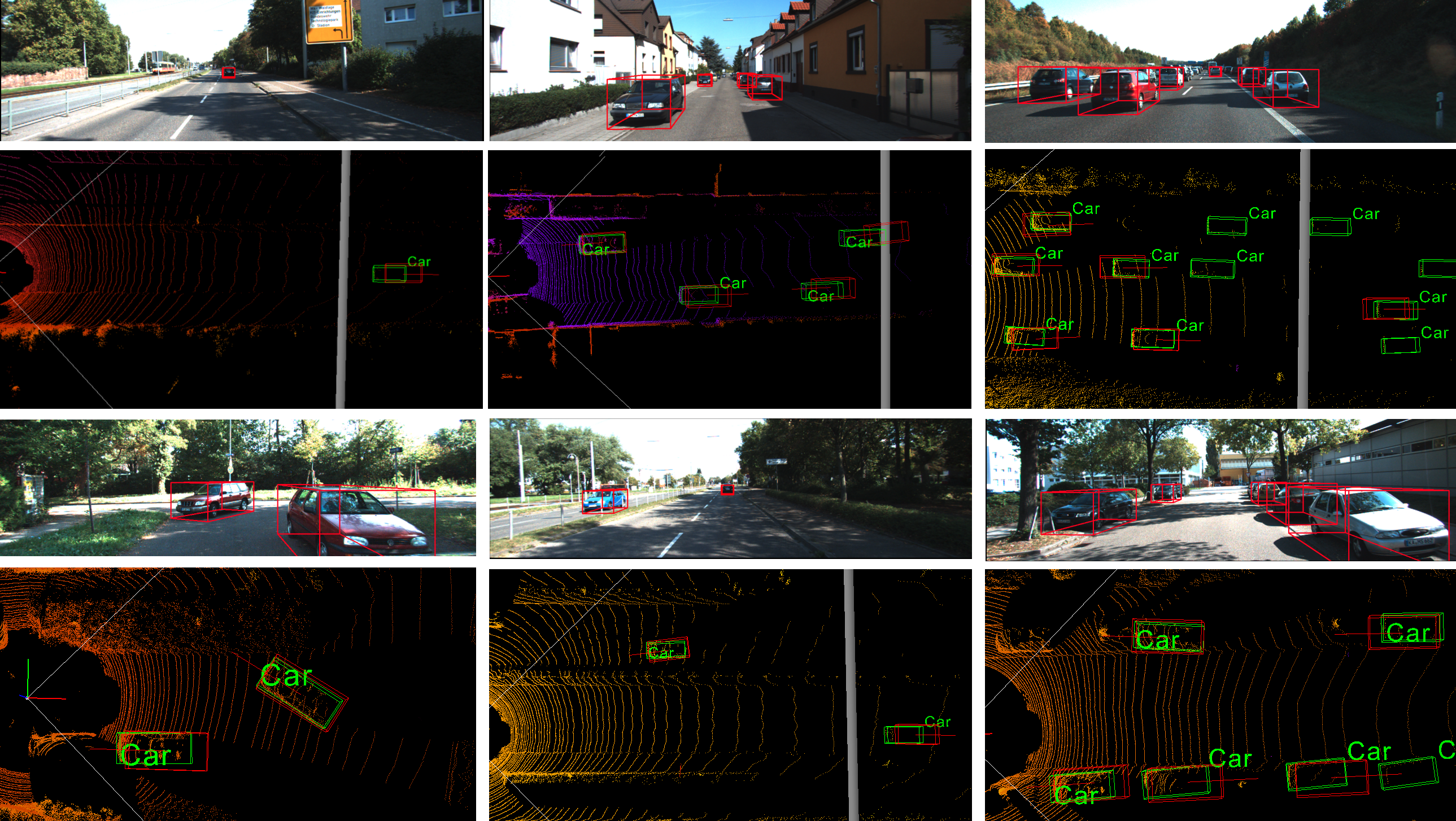}
      
      \caption{Examples of detections (red boxes) generated by MonoFlex trained on our pseudo-labels (best viewed in color)}
      \label{fig:val_qual}
  \end{figure*}

In Table \ref{table:pseudo_labels}, we show the quality of our pseudo-labels on the train split after each iteration. Despite the large domain gap between CARLA and KITTI, and the poor initial estimates, results demonstrate that the pseudo-labels improve as the model is finetuned and produces better initial estimates in the subsequent iterations, surpassing pseudo-labels generated by Autolabeling \cite{autolabeling} (which even uses ground truth boxes) in the AP BEV metric, whilst also achieving a decent 2D accuracy (AP 2D). Figure \ref{fig:pseudo_qual} visualizes our improving pseudo-labels of the train split across the three iterations, while Figure
\ref{fig:val_qual} depicts the inference results on the validation set achieved by MonoFlex trained on our pseudo-labels.
\begin{table}[h!]
\centering
\small
\begin{tabular}{*8c}
 \hline
  Iteration&\multicolumn{3}{c}{AP 2D \%} & \multicolumn{3}{c}{AP BEV \%}\\
 \hline\hline
&Easy&Mod&Hard&Easy&Mod&Hard\\ 
1&84.5&63.2&56.0&66.7&45.0&37.9\\
2&91.5&67.3&57.6&87.2&60.5&50.8\\
3&91.9&69.8&60.1&\textbf{89.9}&\textbf{63.1}&\textbf{53.4}\\
\hline
Autolabeling \cite{autolabeling}& \multicolumn{3}{c}{Ground truth boxes}&77.8&59.7&N/A\\
\hline\hline
\end{tabular}
\caption{Evaluation of the pseudo labels generated during the three iterations that we perform on the train split}

\label{table:pseudo_labels}
\end{table}
  \begin{table*}[h!]
\centering
\small
\def\svgwidth{\textwidth}

\begin{tabular}{l*{8}{c}}
\hline
  
  &&\multicolumn{3}{c}{ $AP_{BEV}$ / $AP_{3D}$ ($AP_{R11}$@ 0.5 IoU) }&\multicolumn{3}{c}{ $AP_{BEV}$ / $AP_{3D}$ ($AP_{R40}$@ 0.5 IoU)}\\
  Method&Images&Easy&Mod&Hard&Easy&Mod&Hard\\
 \hline
 \multicolumn{8}{c}{Supervised}\\
 \hline
Deep3DBBox \cite {mousavian20173d}&trainsplit&30.02/27.04&23.77/20.55&18.83/15.88&-&-&-\\
Mono3D \cite{mono3d} &trainsplit&30.50/25.19&22.39/18.20&19.16/15.52&-&-&-\\
M3D-RPN  \cite{m3drpn} &trainsplit&55.37/48.96&42.49/39.57&35.29/33.01&53.35/48.53&39.60/35.94&31.76/28.59\\
MonoGRNet \cite{monogrnet}&trainsplit&-&-&-&48.53/47.59&35.94/32.28&28.59/25.50\\
LPCG-M3D-RPN \cite{lpcg}&trainsplit&67.66/61.75&\textbf{52.27}/49.51&46.65/\textbf{44.70}&-&-&-\\
MonoFlex \cite{monoflex}&trainsplit&\textbf{68.62}/\textbf{65.33}&51.61/\textbf{49.54}&\textbf{49.73}/43.04&\textbf{67.08}/\textbf{61.66}&\textbf{50.54}/\textbf{46.98}&\textbf{45.78}/\textbf{41.38}\\
\hline
\multicolumn{8}{c}{Unsupervised}\\
\hline
MonoDIS- SDFLabel \cite{autolabeling}& trainsplit&51.10/32.90&34.50/22.10&-&-&-&-\\
\textbf{Ours} w/ MonoFlex &trainsplit&52.43/36.71&37.55/26.74&31.21/22.09&48.59/32.10&31.45/21.12&24.40/15.92\\
MonoDR \cite{monodr}&-&51.13/45.76&37.29/32.31&30.20/26.19&48.53/43.37&33.90/29.50&25.85/22.72\\

LPCG-M3D-RPN\cite{lpcg}&Raw data&52.06/47.58&35.37/29.06&28.61/26.58&-&-&-\\
\textbf{Ours} w/ MonoFlex
&Raw data&\underline{63.94}/\underline{51.90}&\underline{42.29}/\underline{33.24}&\underline{35.31}/\underline{30.39}&\underline{59.63}/\underline{46.95}&\underline{38.31}/\underline{30.08}&\underline{30.62}/\underline{24.41}\\
\hline
 \end{tabular}
\caption{Average precision on KITTI validation set of our method and other supervised and unsupervised methods (best is in bold, best unsupervised is underlined)}

\label{table:val_11}
\end{table*}
\subsection{Comparison with state-of-the-art}
After the end of the third iteration, we evaluate our model on the validation split to compare with the state-of-the-art. Our results from Table \ref{table:val_11} demonstrate the ability of our pseudo-labels to achieve competitive results without using any supervision, even when only training on the train split.

As we can learn without labels, we can easily make use of a huge amount of data to further boost our performance. To demonstrate the usefulness of this, we generate pseudo-labels for KITTI Raw sequences, which are video sequences acquired with synchronized lidar point clouds but without any labeling.
After discarding videos overlapping with the train or validation split, we generate pseudo-labels for 20 sequences and add them to the labels pool created for the train split. We again finetune the model trained on synthetic data on this larger set of sequences, using as supervision the generated pseudo-labels, for three iterations. Results reported in Table \ref{table:val_11} demonstrate the capacity of our pipeline to improve when more data is available outperforming other unsupervised methods, and matching performance obtained by recently proposed fully supervised methods (such as \cite{m3drpn}), underlining the advantage of unsupervised approaches which require no labeling efforts.

\section{Ablation Study}
We first want to demonstrate the impact of our main contribution, being the use of temporal information. To this end, we test our pseudo-label generation without using any temporal prior, but instead solely employing the Chamfer distance for self-supervision.
In Table \ref{table:notemp}, we demonstrate the respective results for pseudo-labels generated by our pipeline, and pseudo-labels generated by the Chamfer baseline, and compare both with the initial model estimates. Our proposed temporal consistency proves to be less sensitive to noisy initial estimates, and maintains a considerable advantage both in AP3D ( Easy 45.45\% compared to 38.61) and AP BEV (Easy 89.90\% compared to 87.23).
\begin{table}[h!]
\centering
\small
\begin{tabular}{*7c}
 \hline
  &\multicolumn{3}{c}{AP 3D \% @0.5IoU} & \multicolumn{3}{c}{AP BEV \% @0.5IoU}\\
 \hline\hline
&Easy&Mod&Hard&Easy&Mod&Hard\\ 
Initals&15.12&9.91&8.18&35.51&21.95&19.70\\
\hline
\multicolumn{7}{c}{1st Iteration}\\
\hline
baseline&20.09&11.72&9.62&54.40&33.00&28.26\\
Ours &33.82&18.78&16.90&66.70&45.02&37.92\\
\hline
\multicolumn{7}{c}{2nd Iteration}\\
\hline
baseline&42.62&25.07&20.28&84.64&55.46&48.23\\
Ours &45.13&29.44&24.54&87.19&60.51&50.82\\
\hline
\multicolumn{7}{c}{3rd Iteration}\\
\hline
baseline&38.61&22.90&20.24&87.23&60.54&50.86\\
Ours &\textbf{45.45}&\textbf{29.38}&\textbf{25.94}&\textbf{89.90}&\textbf{63.12}&\textbf{53.40}\\

\hline

\end{tabular}
\caption{Quantitative evaluation of the pseudo-labels generated for the train split using the Chamfer baseline and our proposed temporal consistency}

\label{table:notemp}
\end{table}

 \subsection{Motion trajectory modeling}
 Modeling the motion of other cars plays an important role in generating accurate pseudo-labels. Noisy initial estimates can complicate the optimization procedure and converge to sub-optimal poses. Using priors on plausible movements of vehicles can potentially alleviate this issue and suppresses noise caused by motion blur or occlusions.
 We test different motion trajectory models, with different complexities and modeling power. First, we apply robust linear fitting (using RANSAC) to the entire trajectory, which assumes that the entire motion is linear within the tracking period. While this model applies to many situations in straight roads, it clearly fails to adapt to other cases such as turns and crossings. As a more complex model, we also consider splines, piece-wise polynomial functions, with an adaptive degree selection (based on MSE between the candidate fit and the original data). Such models, however, require, in addition to the polynomial degree, an adequate smoothing parameter selection (related to the number of knots), which is hard to tune given different motion patterns.
 Our experiment shows that simple models (like the linear one) are still expressive when applied on local segments of the trajectory, while requiring less parameter tuning. 
 
 In Table  \ref{table:motion_models}, we generate the first iteration of pseudo-labels of the train split using different motion models, and we evaluate the moving cars against the ground-truth (since these are of main interest for this ablation).
 \begin{table}[h!]
\centering
\small
\begin{tabular}{*7c}
 \hline
  &\multicolumn{3}{c}{AP 3D \% @0.5IoU} & \multicolumn{3}{c}{AP BEV \% @0.5IoU}\\
 \hline\hline
&Easy&Mod&Hard&Easy&Mod&Hard\\ 
\hline
\multicolumn{7}{c}{Moving}\\
\hline
linear&32.65&21.08&19.54&60.68&45.80&43.36\\
ada. spline &37.92&23.96&22.22&68.23&50.71&45.90\\
p.w linear &\textbf{41.40}&\textbf{26.41}&\textbf{24.58}&\textbf{68.25}&\textbf{50.92}&\textbf{48.47}\\
\hline

\end{tabular}
\caption{Evaluation of moving cars in the pseudo-labels generated using different motion models: global robust linear fit (using RANSAC)- first row, adaptive spline fitting- second row, piece-wise linear fit (using RANSAC)- third row}

\label{table:motion_models}
\end{table}
\subsection{Motion state classification}
The proper use of the temporal prior requires an accurate understanding of the state of the object (in motion, static). By discarding the temporal context of short lived tracklets, we avoid wrongly assigned motion classes and maintain high precision for both motion states. In Table \ref{table:classification}, we compare our tracklets with the ground-truth (of the trainsplit) and measure the precision of our motion classification module across different iterations. We observe a consistent improvement of both the precision and the support, which is achieved by an increasing detection accuracy. Additionally, this is reflected in the support of short tracks due to better matching.

To further analyze the contribution of each of the temporal priors (static and moving objects), we perform an evaluation of our pseudo-labels considering only one motion class at a time and compare with the baseline approach. 
Table \ref{table:stratified} demonstrates the effectiveness of the temporal consistency in both of the motion states, with a stronger improvement in the case of static cars.
\begin{table}[h!]
\centering
\small
\begin{tabular}{*6c}
 \hline
  Iteration&\multicolumn{2}{c}{Static} & \multicolumn{2}{c}{Moving}& 
  undecided \\
 \hline\hline
&P \%$\uparrow$&\#boxes$\uparrow$&P \%$\uparrow$&\#boxes$\uparrow$&\#boxes$\downarrow$\\ 
1st&93.97&3291&96.37&3257&175\\
2nd&99.11&3157&91.85&4129&118\\
3rd &\textbf{99.68}&\textbf{3730}&\textbf{96.40}&\textbf{4369}&\textbf{78}\\
\hline

\end{tabular}
\caption{Classification of object motion state in different iterations of pseudo-labels.}

\label{table:classification}
\end{table}
\begin{table}[h!]
\centering
\small
\begin{tabular}{*7c}
 \hline
  &\multicolumn{3}{c}{AP 3D \% @0.5IoU} & \multicolumn{3}{c}{AP BEV \% @0.5IoU}\\
 \hline\hline
&Easy&Mod&Hard&Easy&Mod&Hard\\ 

\multicolumn{7}{c}{Static}\\
\hline
baseline&7.79&5.31&4.32&39.78&20.43&17.95\\
Ours &\textbf{23.28}&\textbf{14.84}&\textbf{12.57}&\textbf{57.76}&\textbf{35.89}&\textbf{29.29}\\
\hline
\multicolumn{7}{c}{Moving}\\
\hline
baseline&31.63&20.19&18.64&60.79&45.83&43.38\\
Ours &\textbf{41.40}&\textbf{26.41}&\textbf{24.58}&\textbf{68.25}&\textbf{50.92}&\textbf{48.47}\\
\hline

\end{tabular}
\caption{Evaluation results by motion class of both our proposed method and the baseline (computed on the trainsplit)}

\label{table:stratified}
\end{table}

  \section{Conclusions}
We have proposed a self-supervised framework to train monocular 3D object detection methods without the use of ground truth labels. Starting off by an initial training on simple synthetic data, we then use the noisy initial pose and shape estimates to formulate our self-supervised loss terms. Assuming the availability of lidar point clouds during training, and harnessing temporal prior in video sequences, we optimize the model estimates and derive high-quality pseudo-labels. Using these in a finetuning procedure, we obtain results competitive with the state-of-the-art and demonstrate the potential of using large-scale un-annotated data in pushing the model performance.
In the future, the pipeline can be extended to support less accurate depth data coming from stereo-pairs or monocular depth estimation.
{\small
\bibliographystyle{IEEEtran}
\bibliography{root}
}
\end{document}